\documentclass[10pt,twocolumn,letterpaper]{article}

\usepackage{multirow}
\usepackage{colortbl}
\usepackage{graphicx}
\usepackage{algorithm}
\usepackage{algpseudocode}

\usepackage{cvpr}              

\definecolor{cvprblue}{rgb}{0.21,0.49,0.74}
\usepackage[pagebackref,breaklinks,colorlinks,allcolors=cvprblue]{hyperref}

\def\paperID{*****} 
\def\confName{CVPR}
\def\confYear{2025}

\title{Video Set Distillation: Information Diversification and Temporal Densification}

\author{
 \textbf{Yinjie Zhao$^{1,2,3}$, Heng Zhao$^{1,2,4}$, Bihan Wen$^{3}$, Yew-Soon Ong$^{1,2,4}$, Joey Tianyi Zhou$^{1,2}$}\\
$^1$CFAR, Agency for Science, Technology and Research (A*STAR), Singapore\\
$^2$IHPC, Agency for Science, Technology and Research (A*STAR), Singapore\\
$^3$School of EEE, Nanyang Technological University, Singapore\\
$^4$CCDS, Nanyang Technological University, Singapore\\
}

\begin{document}
\maketitle
\begin{abstract}

The rapid development of AI models has led to a growing emphasis on enhancing their capabilities for complex input data such as videos. While large-scale video datasets have been introduced to support this growth, the unique challenges of reducing redundancies in video \textbf{sets} have not been explored. Compared to image datasets or individual videos, video \textbf{sets} have a two-layer nested structure, where the outer layer is the collection of individual videos, and the inner layer contains the correlations among frame-level data points to provide temporal information. Video \textbf{sets} have two dimensions of redundancies: within-sample and inter-sample redundancies. Existing methods like key frame selection, dataset pruning or dataset distillation are not addressing the unique challenge of video sets since they aimed at reducing redundancies in only one of the dimensions. In this work, we are the first to study Video Set Distillation, which synthesizes optimized video data by jointly addressing within-sample and inter-sample redundancies. Our Information Diversification and Temporal Densification (IDTD) method jointly reduces redundancies across both dimensions. This is achieved through a Feature Pool and Feature Selectors mechanism to preserve inter-sample diversity, alongside a Temporal Fusor that maintains temporal information density within synthesized videos. Our method achieves state-of-the-art results in Video Dataset Distillation, paving the way for more effective redundancy reduction and efficient AI model training on video datasets.

\end{abstract}    
\section{Introduction}
\label{sec:intro}

\begin{figure}[ht]
    \centering
    \includegraphics[width=1.2\columnwidth, trim=140 35 50 50]{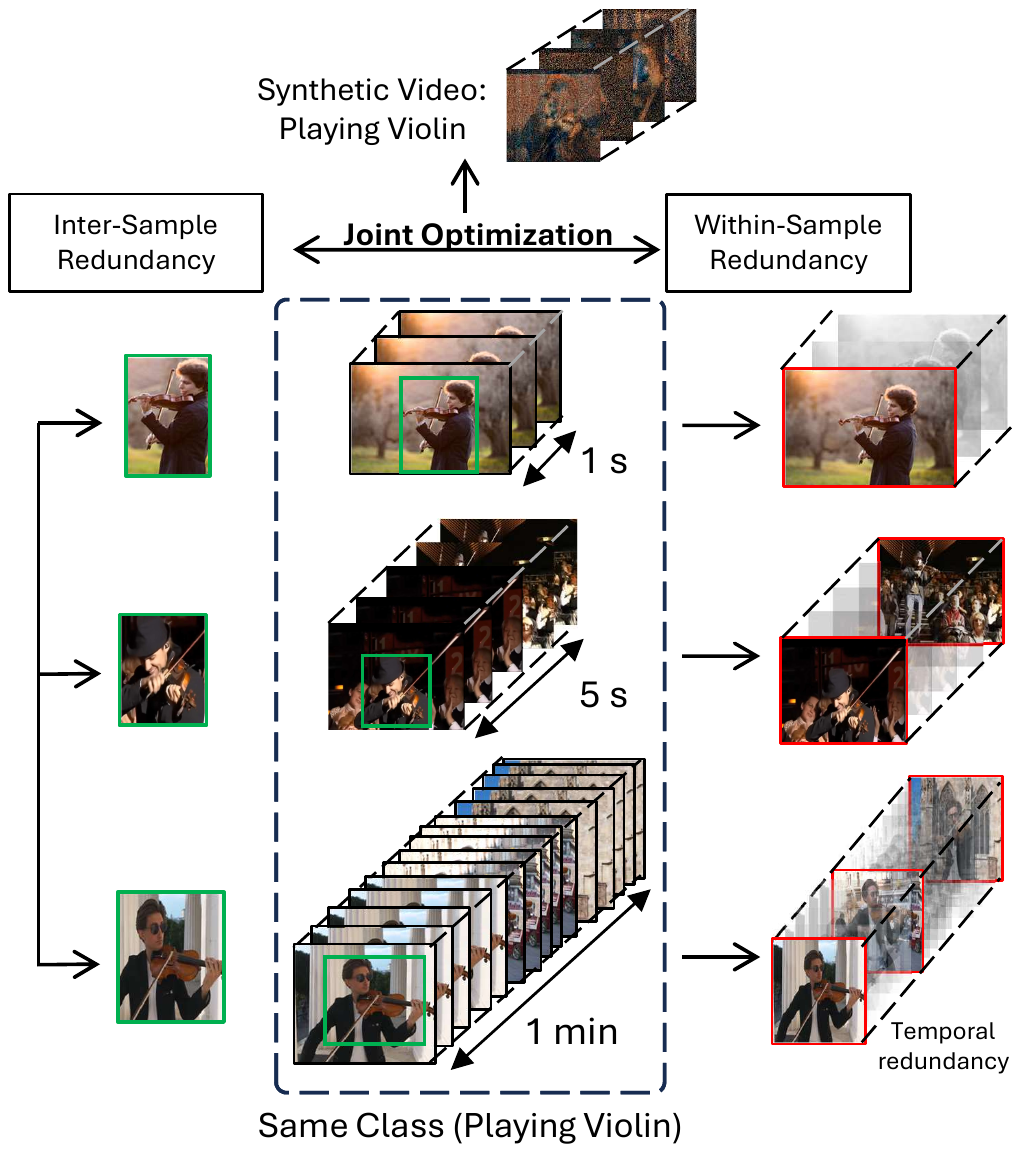}
    \caption{The grand challenge of Video Set Distillation comes from its nature as a two-layer nested set. Each individual video is a set of image-level data points, and video set is a set of individual videos. It is critical to jointly optimize over both inter-sample redundancy and the within-sample redundancy. Furthermore, given the large variety of temporal length in a video set, the within-sample redundancies are largely different from each other, increasing the complexity of redundancy reduction. }

    \label{fig:task_cha}
\end{figure}

With the rapid advancement of AI models, there has been an increasing focus on enhancing AI capabilities for complex input data such as videos. This also encourages the emergence of large scale video datasets proposed to accelerate the development of AI in this domain \cite{dst1, dst2, dst3, vidrec6}. However, compared to other data modality such as text and image, video data consumes significantly more storage and also requires more computational power to analyze. How to effectively reduce the size of this ever-growing pool of video datasets with a focus on training and evaluation of multi-modal machine intelligence have become an critical and practical problem. 

It is apparent that video data contains a high level of redundancy in terms of information density. We identify two dimensions of redundancy that could be examined to reduce the size of a video dataset: within-sample redundancy(similar frames in a same video); b) inter-sample redundancy(similar videos in a same dataset). There are many approaches that could be applied to these two redundancy dimensions; such as Key Frame Selection\cite{naive_met1} or down-sampling for within-sample redundancy and Dataset Pruning \cite{naive_met2} for inter-sample redundancy. However, these naive methods are basically pruning the information by keeping some information intact and throwing the other away without considering how AI model will perform on the pruned data. Alternatively in this paper, we study a dataset synthesis problem named Video Set Distillation which aims to generate video datasets that are much smaller than the original by jointly optimizing within-sample, inter-sample redundancy and AI model performance. Specifically, Video Set Distillation learns to project similar raw video frames into a reduced number of synthesized frames within one video sample; and at the same time learn to combine multiple similar video clips into one synthesized sample via generation process.

It is worth noting that Video Set Distillation is different from the traditional research domain of video compression which primarily aims to reduce the size of individual video sample without altering their visual quality and temporal consistency in ways perceptible to humans. In contrast, Video Set Distillation serves a machine-centric purpose, aiming to generate a much smaller synthetic dataset optimized for efficient model training and evaluation, with minimal concern for human interpretability of the synthetic data.



Due to the nature of the input video data, Video Set Distillation needs to tackle two dimensions of redundancy: within-sample redundancy and inter-sample redundancy. Specifically, within-sample redundancy refers to the existence of similar frames in a same video clip and similar pixels in a same certain frame while inter-sample redundancy refers to the existence of highly similar video clips in one video dataset, as shown in Fig.\ref{fig:task_cha}. Although there are existing studies on Dataset Distillation (DD)\cite{dd1} which generates synthetic data via optimization processes, they only focus on image which is a simple form of data and hence cannot be applied directly for Video Set Distillation. Different from images, videos contain action related semantics that are embedded in temporal sequences; and the information density in each video clip sample is different. For example in Fig.\ref{fig:task_cha}, the video samples for one action can have lengths ranging from 1 second to 1 minute, but they all share the same class of "playing violin". One challenge in Video Set Distillation is how to reduce the temporal redundancy for video samples that have different information density automatically and adaptively. Another challenge is how to effectively reduce the inter-sample redundancy within a same class and temporal redundancy within a same video clip simultaneously.

Existing work such as VDSD\cite{dd12} tried to tackle individual video data by simply condensing video clip into a single static image, ignoring the rich temporal information. To address this issue, we propose Information Diversification and Temporal Densification (\textbf{IDTD}), which aims to jointly reduce both redundancies in an end-to-end manner. Firstly for reducing inter-sample redundancy, we enhance information diversification by leveraging a Feature Pool shared by multiple Feature Selectors, where the information diversity is achieved by Feature Selectors driven by a diversity loss function, while the common knowledge within the class is preserved by the Feature Pool. To efficiently utilize the limited Instance Per Class (IPC) budget for each class, it is of crucial importance to more comprehensively represent the information distribution from the original dataset. Secondly, to reduce within-sample redundancy; the second part of our mechanism is a Temporal Fusor seamlessly integrating the diverse information into the synthetic video. It takes the diverse feature generated by the first part as input and generates the synthetic videos. Such process enables the feature diversity to exist along the temporal dimension, and ensures the temporal information density since it forces the synthetic video to maintain diverse feature in a limited temporal size.


Our contribution can be summarized as:
\begin{itemize}
    \item We are the first to identify and tackle the two dimension of redundancy in Video Set Distillation by designing specific module for each type of redundancy.
    \item We propose IDTD to jointly optimize over within-sample and inter-sample redundancies on video \textbf{sets} in an end-to-end manner. We also propose an information diversification module to reduce the inter-sample redundancy while at the same time maximize the feature diversity of the generated synthetic video sample. 
    \item Comprehensive experiments are conducted on multiple datasets, the results demonstrate the superiority of our proposed approach which achieves SOTA in Video Set Distillation on all datasets.

\end{itemize}

\begin{figure*}[ht]
    
    \centering
    \includegraphics[width=0.7\textwidth,trim = 130 170 130 200]{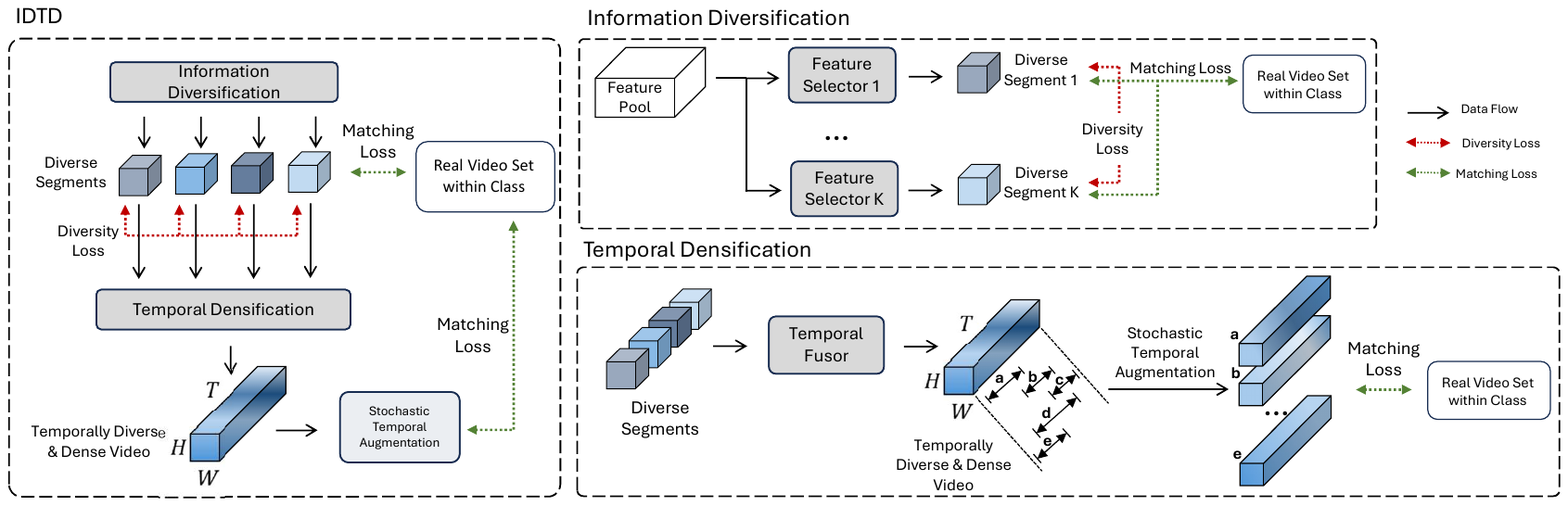}
    \caption{As shown on the left, Our IDTD approach jointly conduct the Information Diversification and the Temporal Densification in an end-to-end manner. On the top right, the Information Diversification part is shown in detail. The Feature Pool is a learnable variable and the Feature Selectors are learnable modules. On the bottom right, the Temporal Diversificaiton part is shown in detail. The Temporal Fusor is a learnable module. The Diversity loss enforces $K$ diverse segments to represent information from the original dataset, and the Temporal Fusor is optimized to integrate the diverse feature into a synthetic video instance driven by a dataset distillation matching loss. $K$ is a hyperparameter determining number of Feature Selectors for each synthetic instance. Our approach effectively balance the redundancy reduction between inter-sample dimension and within-sample dimension, while it keeps the temporal diverse information.}
    \label{fig:idtd_diagram}
    
\end{figure*}

\section{Related Works}
\label{sec:related_works}

\subsection{Dataset Distillation}

Existing Dataset Distillation (DD)\cite{dd1} approaches have primarily focused on image-level dataset distillations. They aims at reducing inter-sample redundancy by synthesizing a compact dataset of a limited Instance Per Class (IPC). Approaches such as \cite{dd3, dd2, dd11} adopt a bi-level optimization perspective and align the training dynamics between the synthetic dataset and the real dataset. Alternatively, approaches like \cite{dd7, dd10, dd13} aligns the features extracted between the synthetic and the real dataset. However, these approaches fails to explicitly consider within-sample redundancy of the dataset, which is crucial for video datasets due their temporal redundancy. Consequently, these approaches' performance remains suboptimal in redundancy reduction for video sets.

Several DD approaches were developed to handle larger scale image datasets. At a larger scale, the redundancy of the dataset increases significantly as scale grows, which is observed similar as in video datasets. TESLA\cite{dd4} addresses the memory consumption issue of large scale datasets, and utilized soft labels to encode more information into the synthetic dataset. However, such memory efficiency has a questionable generalizability to large scale video datasets. Redundancy in video sets increases at a much faster rate as the growth of sample size due to the additional of temporal dimension. SRe$^{2}$L\cite{dd4_1} and RDED\cite{Diversity_4} reduce computational cost by leveraging the knowledge from well-trained teacher model instead of training the student model in the bi-level optimization process. Nonetheless, these approaches are limited to image-level datasets and failed to addressed the issue of temporal redundancy reduction, leaving the key challenge of redundancy reduction in video sets unresolved.

Recent works have sought to extend DD approaches beyond image-level datasets. Text dataset distillation has been explored by \cite{dd5}, and \cite{dd8} explored multi-modality dataset distillation with both image and text modalities. VDSD\cite{dd12} tried to adapt DD techniques on video datasets. However, the approach treats the contribution of temporal information as trivial. They distilled a static image from the real dataset and compensated it with dynamic information using a trained interpolator. As a result, VDSD has a limited improvement on video datasets compared to existing image-level DD approaches.

\subsection{Video Recognition and Redundancy}

Video Recognition was firstly explored in the deep learning context by \cite{vidrec1}. Convolutional Networks such as C3D\cite{vidrec5} and I3D\cite{vidrec6} paved ways for video feature extraction foundation models with deep learning. Since the early days of video recognition studies, redundancies in videos have been a critical factor influencing task effectiveness. For example, \cite{vidrec3} splits the feature extraction process into appearance and motion pathways, leveraging residual networks\cite{vidrec4} to model the interaction between these pathways. Such approach jointly reduces video redundancy in an implicit way by addressing the interaction between two specific pathways. SlowFast\cite{vidrec2} explicitly considers the temporal redundancy of video modality, employing two different temporal granularity of temporal information extraction. Such redundancy reduction endeavor underscore the importance of redundancy reduction in video data without compromising valuable information.


This principle of redundancy reduction at the model level is equally applicable when aiming to reduce redundancy in video sets themselves. However, existing research has not adequately addressed the complexity of video redundancies. A comprehensive approach should jointly consider both the inter-sample and within-sample redundancies, ensuring effective redundancy reduction in video sets.

\section{Methodology}

\subsection{IDTD}

Most of existing works did not explicitly address the influence of temporal redundancy and information diversity to a video \textbf{set} distillation process. To tackle this issue, we propose a novel approach that leveraged a joint-optimization framework as outlined in the following sections as illustrated by Fig.\ref{fig:idtd_diagram}.

The problem of Video Set Distillation on a real video dataset $\mathcal{T}$ can be formulated as follows: Given $N$ classes and $M$ Instance Per Class (IPC), the objective is to generate a synthetic video set $V_{syn} = \{\{v_{m,n}\}_{m=1}^{M}\}_{n=1}^{N}$, where $v_{m,n} \in \mathbb{R}^{H \times W \times C \times T}$.

The effectiveness of existing DD approaches is significantly compromised when applied to video datasets due to the interplay between within-sample redundancy and inter-sample redundancy. To address this limitation, we propose an approach that jointly promotes higher information diversity while reducing temporal redundancy.

As illustrated in Fig.\ref{fig:idtd_diagram}, our IDTD approach comprises two primary components: the diversification module and the fusion module. The diversification module is designed to generate segments with diverse features, optimized jointly through a diversity loss and a dataset distillation loss. Subsequently, the Temporal Fusor takes these diverse segments as input and synthesizes the final video. 

Compared to existing approaches, our IDTD approach explicitly encourage diversity, which enhances the effectiveness and information density of the synthetic dataset.

\begin{algorithm}
\caption{$IDTD(\mathcal{T}) \to V_{syn}$}
\label{alg:algo_chart}
\begin{algorithmic}[1]
\State \textbf{Input:} Training Set $\mathcal{T}$
\State \textbf{Output:} Synthetic Video Set $V_{syn}$ 
\State Initialize variables Feature Pool $P$, Feature Selector $S$, Temporal Fusor $F$; Hyperparameter $K$; Number of Classes $N$, $IPC=M$\\

\State \textbf{Training}
\For{it in Iterations}
    \For{$n$ in $N$}
        \For{$m$ in $M$} 
            \For{$k$ in $K$}
                \State Produce Diverse Segments:
                \State $d_{m,n,k} = s_{m,n,k}(p_{m,n})$
            \EndFor
            \State $ v_{m,n} = f_{m,n}(\{d_{m,n,k}\}_{k=1}^{K})$
            \State $v'_{m,n} = \tau (v_{m,n})$
        \EndFor
        \State Calculate loss by equation \ref{eq:loss_eq}.
        \State Update $P$, $S$, $F$
    \EndFor
\EndFor
\\
\State \textbf{Synthesization}
\For{$n$ in $N$}
    \For{$m$ in $M$}
        \State $ v_{m,n} = f_{m,n}(\{s_{m,n,k}(p_{m,n})\}_{k=1}^{K})$
    \EndFor
\EndFor
\\
\State \textbf{return} $V_{syn}$

\end{algorithmic}
\end{algorithm}

\subsection{Information Diversification}
\label{info_div}

The information diversification process consists of a Feature Pool and a set of Feature Selectors. Each Feature Selector takes the Feature Pool as input and generates a video segment. The generated video segments are explicitly enforced to have distinct features. At the same time, because all Feature Selectors share the same Feature Pool as input, this structure ensures that the Feature Pool gathers meaningful information, regardless of how diverse the segments are.

The Features Pool can be expressed as $P = \{\{p_{m,n}\}_{m=1}^{M}\}_{n=1}^{N}$ and Feature Selectors $S=\{\{\{s_{m,n,k}\}_{k=1}^{K}\}_{m=1}^{M}\}_{n=1}^{N}$, where $K$ is a hyperparameter that determines the number of Feature Selectors and, consequently, the number of feature Segments per instance. The set of feature segments is collectively referred to as the Diverse Segments $D=\{\{\{d_{m,n,k}\}_{k=1}^{K}\}_{m=1}^{M}\}_{n=1}^{N}$.

The process of producing the Diverse Segments can be expressed as:
$$d_{m,n,k} = s_{m,n,k}(p_{m,n})$$

To ensure diversity among the information contained in the Diverse Segments, a diversity loss is applied. It is defined as the L2 distance between each pair of distinct Diverse Segments:
\begin{equation}
    \mathcal{L}_{div}=\sum_{k=1}^{K-2} \sum_{q=k+1}^{K} || (\phi(d_{m,n,k}) - \phi(d_{m,n,q})||_{2}^{2}
    \label{eq:loss_eq}
\end{equation}

where $\phi$ is the feature extraction process of the student model's feature extraction process.

Simultaneously, the Diverse Segments are also optimized by dataset distillation matching loss between the Diverse Segments $D$ and the real dataset $\mathcal{T}$, expressed as $\mathcal{L}_M(D,\mathcal{T})$.

\subsection{Temporal Densification}

After $D$ is computed, the Temporal Fusor integrates the segments in the temporal dimension, defined as $ F = \{\{f_{m,n}\}_{m=1}^{M}\}_{n=1}^{N}$. The input of the Temporal Fusor is the Diverse Segments, and the output is the synthetic video of target size. The Temporal Densification process can be expressed as:
$$ v_{m,n} = f_{m,n}(\{d_{m,n,k}\}_{k=1}^{K})$$

However, once $v_{m,n}$ is generated, the temporal sequence and absolute positions of information within the synthetic video become fixed. This rigidity significantly limits the utility of information diversity achieved through \ref{info_div}. To address this limitation, we introduce Stochastic Temporal Augmentation, analogous to data augmentation in image-level trainings. This augmentation randomly samples a temporal fraction of $v_{m,n}$ and reshape it to the targeted temporal size of the synthetic video. The process can be described as:
$$v'_{m,n} = \tau (v_{m,n}, \mu)$$
$$\mu = [s,e], \quad s<e \quad and \quad s,e\in[0,1]$$

where $\tau(\cdot)$ is the temporal augmentation process, and $\mu$ is a random temporal interval. The synthetic dataset after applying the Stochastic Temporal Augmentation is defined as $V'_{syn} = \{\{v'_{m,n}\}_{m=1}^{M}\}_{n=1}^{N}$ 

\subsection{Training Objectives}

Our approach seamlessly integrate the objectives of information diversification and temporal densification with existing dataset distillation losses. Specifically, we apply a DD matching loss to both the Diverse Segments and the final synthetic video, and meanwhile a diversification loss is applied among the Diverse Segments. These losses are optimized jointly to achieve an effective balance between diversity and representativeness. Given a dataset distillation matching objective function between synthetic and real videos $\mathcal{L}_M(\cdot, \cdot)$, the overall training objective is defined as:
$$\mathcal{L} = \mathcal{L}_M((D, \mathcal{T}) + \alpha_1 * \mathcal{L}_{div} + \alpha_2*\mathcal{L}_M((V'_{syn}, \mathcal{T})$$
where $\alpha_1$ and $\alpha_2$ are hyperparameters determining the contribution of each loss.

This training objective encourages Diverse Segments to diverge from each other (via $\mathcal{L}_{div}$), while simultaneously aligning the synthetic features to the real data (via $\mathcal{L}_M$). By doing so, the within-sample redundancy and inter-class redundancy are jointly reduced, enabling the synthetic video set to capture more meaningful and representative information.

To elaborate on the training dynamics,  the Diverse Segments $\{d_{m,n,k}\}_{k=1}^{K}$ share a common Feature Pool and are collectively optimized using the dataset distillation matching loss. This combination of losses forces the Feature Pool to encapsulate the essential knowledge within each class, and the Feature Selectors learns to generate diverse information within the same class. 

\subsection{Training Pipeline}

As shown in Alg.\ref{alg:algo_chart}, the training process and synthesization process of our approach share the same data flow, with the only difference of Stochastic Temporal Augmentation applied and calculating the overall loss with equation \ref{eq:loss_eq}.

Our training pipeline updates the Feature Pool, the $K$ Feature Selectors and the Temporal Fusor on a class-by-class basis each optimization iteration. Such updating mechanism reduces the memory consumption compared to updating all classes by one backward propagation within an iteration, making it computationally efficient. 
\section{Experiment}

\begin{table*}[ht]
\centering
\begin{tabular}{c|c|c|c|c|c}
\hline
\multicolumn{2}{c|}{\textbf{Dataset}} & \multicolumn{2}{c|}{\textbf{MiniUCF}} & \multicolumn{2}{c}{\textbf{HMDB51}} \\
\multicolumn{2}{c|}{\textbf{IPC}} & \multicolumn{1}{c}{1} & \multicolumn{1}{c|}{5} & \multicolumn{1}{c}{1} & \multicolumn{1}{c}{5} \\

\hline

\multicolumn{2}{c|}{\textbf{Full Dataset}} & \multicolumn{2}{c|}{57.22 $\pm$ 0.14} & \multicolumn{2}{c}{28.58 $\pm$ 0.69}\\

\hline

\multicolumn{2}{c|}{ \begin{tabular}{c|c}
    \multirow{3}{*} {Coreset Selection} & Random \cite{dd12} \\
     & Herding \cite{exp1} \\
     & K-Center\cite{exp2}
\end{tabular} }  & 

\multicolumn{2}{c|}{ \begin{tabular}{c c}
    \multicolumn{1}{c}{9.9$\pm$ 0.8} & \multicolumn{1}{c}{22.9$\pm$1.1}\\
   \multicolumn{1}{c}{12.7$\pm$1.6}  & \multicolumn{1}{c}{25.8$\pm$0.3}\\
    \multicolumn{1}{c}{11.5$\pm$0.7} & \multicolumn{1}{c}{23.0$\pm$1.3}
\end{tabular} } 

& 
\multicolumn{2}{c}{ \begin{tabular}{c c}
    \multicolumn{1}{c}{4.6$\pm$0.5} & \multicolumn{1}{c}{6.6$\pm$0.7}\\
   \multicolumn{1}{c}{3.8$\pm$0.2}  & \multicolumn{1}{c}{8.5$\pm$0.4}\\
    \multicolumn{1}{c}{3.1$\pm$0.1} & \multicolumn{1}{c}{5.2$\pm$0.3}
\end{tabular} } 

 \\
 
\hline
\multicolumn{2}{c|}{ \begin{tabular}{c|c}
    \multirow{7}{*} { Dataset
 Distillation } 
     & DM\cite{dd10} \\
     & MTT\cite{dd2} \\
    & FRePo\cite{dd11} \\
     & DM\cite{dd10} + VDSD\cite{dd12} \\
     & MTT\cite{dd2} + VDSD\cite{dd12} \\
     & FRePo\cite{dd11} + VDSD\cite{dd12} \\
     & IDTD (Ours)
\end{tabular} }  & 

\multicolumn{2}{c|}{ \begin{tabular}{c c} 
\multicolumn{1}{c}{15.3$\pm$1.1} & \multicolumn{1}{c}{25.7$\pm$0.2} \\
\multicolumn{1}{c}{19.0$\pm$0.1} & \multicolumn{1}{c}{28.4$\pm$0.7} \\
    \multicolumn{1}{c}{20.3$\pm$0.5} & \multicolumn{1}{c}{30.2$\pm$1.7} \\
    \multicolumn{1}{c}{17.5$\pm$0.1} & \multicolumn{1}{c}{27.2$\pm$0.4} \\
     \multicolumn{1}{c}{\textbf{23.3$\pm$0.6}} & \multicolumn{1}{c}{28.3$\pm$0.0}\\
   \multicolumn{1}{c}{22.0$\pm$1.0}  & \multicolumn{1}{c}{31.2$\pm$0.7}\\
    \multicolumn{1}{c}{22.52$\pm$ 0.1} & \multicolumn{1}{c}{\textbf{33.29$\pm$0.5}}
\end{tabular} } 

& 
\multicolumn{2}{c}{ \begin{tabular}{c c}
    \multicolumn{1}{c}{6.1$\pm$0.2} & \multicolumn{1}{c}{8.0$\pm$0.2} \\
   \multicolumn{1}{c}{6.6$\pm$0.5}  & \multicolumn{1}{c}{8.4$\pm$0.6} \\
    \multicolumn{1}{c}{7.2$\pm$0.8} & \multicolumn{1}{c}{9.6$\pm$0.7} \\
    \multicolumn{1}{c}{6.0$\pm$0.9} & \multicolumn{1}{c}{8.2$\pm$0.4} \\
    \multicolumn{1}{c}{6.5$\pm$0.4} & \multicolumn{1}{c}{8.9$\pm$0.1} \\
    \multicolumn{1}{c}{8.6$\pm$0.1} & \multicolumn{1}{c}{10.3$\pm$0.6} \\
    \multicolumn{1}{c}{\textbf{9.52$\pm$0.3}} & \multicolumn{1}{c}{\textbf{16.15$\pm$0.9}}
\end{tabular} } 

 \\
\hline
\end{tabular}
\caption{Comparison with existing approaches on small scale datasets. We compared with the current main stream image level dataset distillation approaches as well existing video distillation approaches. Our approach obtained a better performance compared to those not considering joint optimization of both redundancies.}
\label{tab:comparison}
\end{table*}

\subsection{Datasets}

We evaluated on commonly used Video recognition datasets such as UCF101\cite{dst1} HMDB51\cite{dst2}, Something-SomethingV2 (SSv2)\cite{dst3} and Kenetics400 (K400)\cite{vidrec6}. For a fair comparison, we evaluated our approach's performance of MiniUCF following VDSD\cite{dd12}, and we divide the datasets into light-weight track (UCF101 \& HMDB51) and heavy-weight track (SSv2 \& K400).

UCF101 includes 101 action classes focusing on human action and interaction with objects. It has a total of 13320 videos, with on average 131 videos per class and average time duration from 2-14 seconds per class. HMDB51 is also a small scale dataset with 6849 videos in total and 51 classes. SSv2 is large scale dataset with 220,847 videos in total with 174 classes. K400 is the most challenging dataset among the datasets, consisting of 400 classes in total. Despite different numbers of classes among these datasets, the duration range of the videos in the datasets are relatively similar, which are mostly within 10 seconds.

\subsection{Implementation Details}

To implement our approach, the Feature Pool is a randomly initialized learnable tensor and the Feature Selectors are randomly initialized light-weighted linear layers. In the Temporal Fusor, concatenated Diverse Segments are input into a light-weighted convolution layer for convolution along the time dimension. For a fair comparison, we used the same student model as VDSD\cite{dd12} which is ConvNet3D. 

We set the learning rate to be 0.01. For the loss weight hyperparameters, we set $\alpha_1 = 0.05$ and $\alpha_2 = {10}^{-4}$ for all datasets. $\alpha_2$ is two levels of magnitude smaller than $\alpha_1$ to prevent gradient explosion at the down stream of the computational graph. We set $K=8$ for all experiments for a reasonable trade-off between training efficiency and performance. For the data loading settings, we followed the same video spatial and temporal size as VDSD\cite{dd12} for a fair comparison.

\subsection{Evaluation Metrics}

We applied the conventional evaluation metrics of Dataset Distillation: given a certain Instance Per Class (IPC) of the synthetic data, we measure the effectiveness of the synthetic dataset based on the evaluation accuracy of the student model on the original test set. When training the student model with the synthetic data, we applied temporal augmentation as we proposed, and train the model for a number of epochs until the loss converges .

\setlength{\tabcolsep}{1pt} 
\begin{table}[ht]
\centering
\begin{tabular}{c|c|c|c|c}
\hline
\multicolumn{1}{c|}{\textbf{Dataset}} & \multicolumn{2}{c|}{\textbf{K400}} & \multicolumn{2}{c}{\textbf{SSv2}} \\
\multicolumn{1}{c|}{\textbf{IPC}} & \multicolumn{1}{c}{1} & \multicolumn{1}{c|}{5} & \multicolumn{1}{c}{1} & \multicolumn{1}{c}{5} \\

\hline

\multicolumn{1}{c|}{\textbf{Full Dataset}} & \multicolumn{2}{c|}{34.6$\pm$0.5} & \multicolumn{2}{c}{29.0$\pm$0.6}\\

\hline
\begin{tabular}{c}
    Random\\
    DM\cite{dd10} \\
     MTT\cite{dd2} \\
      DM\cite{dd10} + VDSD\cite{dd12} \\
      MTT\cite{dd2} + VDSD\cite{dd12} \\
      IDTD (Ours) 
\end{tabular}
       
& 

\multicolumn{2}{c|}{ \begin{tabular}{c c} 
3.0$\pm$0.1 & 5.6$\pm$0.0\\
6.3$\pm$0.0  & 9.1$\pm$0.9 \\
    3.8$\pm$0.2 & 9.1$\pm$0.3 \\
 6.3$\pm$0.2   &7.0$\pm$0.1\\
 \multicolumn{1}{c}{\textbf{6.3$\pm$0.1}}  & 11.5$\pm$0.5\\
   6.1$\pm$0.1  & \multicolumn{1}{c}{\textbf{12.1$\pm$0.2}}\\
\end{tabular} } 

& 
\multicolumn{2}{c}{ \begin{tabular}{c c}
    \multicolumn{1}{c}{3.3$\pm$0.1} & \multicolumn{1}{c}{3.9$\pm$0.1} \\
   \multicolumn{1}{c}{3.6$\pm$0.0}  & \multicolumn{1}{c}{4.1$\pm$0.0} \\
    \multicolumn{1}{c}{3.9$\pm$0.1} & \multicolumn{1}{c}{6.3$\pm$0.3} \\
    \multicolumn{1}{c}{4.0$\pm$0.1} & \multicolumn{1}{c}{3.8$\pm$0.1} \\
    \multicolumn{1}{c}{\textbf{5.5$\pm$0.1}} & \multicolumn{1}{c}{8.3$\pm$0.1} \\
    \multicolumn{1}{c}{3.9$\pm$0.1} & \multicolumn{1}{c}{\textbf{9.5$\pm$0.3}}
\end{tabular} } 

 \\
\hline
\end{tabular}
\caption{Comparison with existing approaches on large scale datasets. Our approach shows advantage at higher IPC. This could because of an increase of information diversity as IPC increases, and lower IPC is more difficult to be diverse enough to produce representative features.}
\label{tab:comparison_2}
\end{table}

\subsection{Comparison with Existing Approaches}

We compared our approaches with existing image level DD approaches as well as approaches targeting the specific challenge of videos.

It is worth noticing that the experiment result of our approach is achieved based on the baseline of Distribution Matching (DM) \cite{dd10}. Other training dynamics matching approaches like MTT\cite{dd2} or FrePo\cite{dd11} take much more computational resources than DM based approach. Therefore, although our performance achieved significant improvements compared to all of other approaches, it would be more fair for our approach to be compared with Distribution Matching based approaches.

For small scale datasets of MiniUCF and HMDB51, it is observed that our approach have larger advantages at higher IPCs. This is explained by our training objectives of Information Diversification. As the IPC increases, there is a higher potential of information diversity that could be utilized and distilled into the synthetic dataset. Furthermore, we noticed that all previous approaches performed poorly on HMDB51 compared to that on MiniUCF. This can be explained that for a more difficult datasets like HMDB51, there are more hard samples or diverse samples, and those samples could be better utilized with an Information Diversification scheme. Although these hard cases lowers the performance upper limit on the full dataset, this becomes an advantage for our approach since we are able to leverage the diversities. 

For large scale datasets including K400 and SSv2, we observed that the performance at low $IPC=1$ is only comparable with existing approaches, while at higher $IPC=5$, the performance starts to show advantage. This could because that as the synthetic dataset scale increases, information diversity becomes more accessible. We can also observe that almost all approaches at a low $IPC=1$ have difficulties surpassing the image-level approaches baselines. This shows that video set distillation at a larger scale remains challenging ans should be further explored in the future works.

\setlength{\tabcolsep}{3pt} 
\begin{table}[ht]
\centering
\begin{tabular}{c|c c|c}
\hline
 &\multicolumn{1}{c}{
     \begin{tabular}{c}
        Feature Pool     \\
        \& Selectors
     \end{tabular}
     } 
     & 
 \multicolumn{1}{c|}{
 \begin{tabular}{c}
        Temporal    \\
         Fusor
     \end{tabular}
  } 
 
 & MiniUCF \\
 
 \hline
 
A &  &  & 14.99$\pm$0.2\\

B  &  & \checkmark  &  18.40$\pm$0.3\\

C & \checkmark &  & 18.57$\pm$0.8\\

IDTD & \checkmark & \checkmark & 22.52$\pm$ 0.1\\

\hline
\end{tabular}
\caption{Ablation on modules. We can interpret the ablation as the following. A: Compress and Stitch; B: Temporal Densification only; C: Information Diversification only.}
\label{tab:ablation1}
\end{table}

\begin{table}[ht]
\centering
\begin{tabular}{c|c c c|c}
\hline
 &\multicolumn{1}{c}{ 
 \begin{tabular}{c}
     Selector \\
    $\mathcal{L}_M$ 
 \end{tabular}

 } &
 \multicolumn{1}{c}{
  \begin{tabular}{c}
     Selector \\
    $\mathcal{L}_{div}$
 \end{tabular}

 } &
 
 \multicolumn{1}{c|}{

   \begin{tabular}{c}
     Fusor \\
     $\mathcal{L}_M$
 \end{tabular}
 } & MiniUCF \\
 
 \hline
A & & & \checkmark & 13.82$\pm$0.3\\

B & & \checkmark & \checkmark & 15.81$\pm$0.6\\

C & \checkmark & & \checkmark & 20.04$\pm$0.4\\

IDTD & \checkmark & \checkmark & \checkmark & 22.52$\pm$ 0.1\\

\hline
\end{tabular}
\caption{Ablation on training objectives. In this table, Selector $\mathcal{L}_M$ is the dataset distillation matching loss applied to the Feature Selectors and Feature Pool, Selector $\mathcal{L}_{div}$ is the diversity loss and Fusor $\mathcal{L}_M$ is the matching loss applied to the Temporal Fusor.
}
\label{tab:ablation2}
\end{table}

\subsection{Ablation Study}

In the ablation study, we aim to explore the contribution of different parts of our approach to the final performance. We mainly evaluate it from three aspects: a) contributions of different modules and their mutual influence; b) contributions of different losses; c) impact of different temporal size of the synthetic videos; d) impact of different level of redundancies from within-sample and inter-sample. Via these ablations, we aim to fully verify the effectiveness of the approach we proposed and how the within-sample and inter-sample redundancies interacts. 

\begin{figure}[ht]
    
    \centering
    \includegraphics[width=0.5\textwidth, trim = 100 70 0 45]{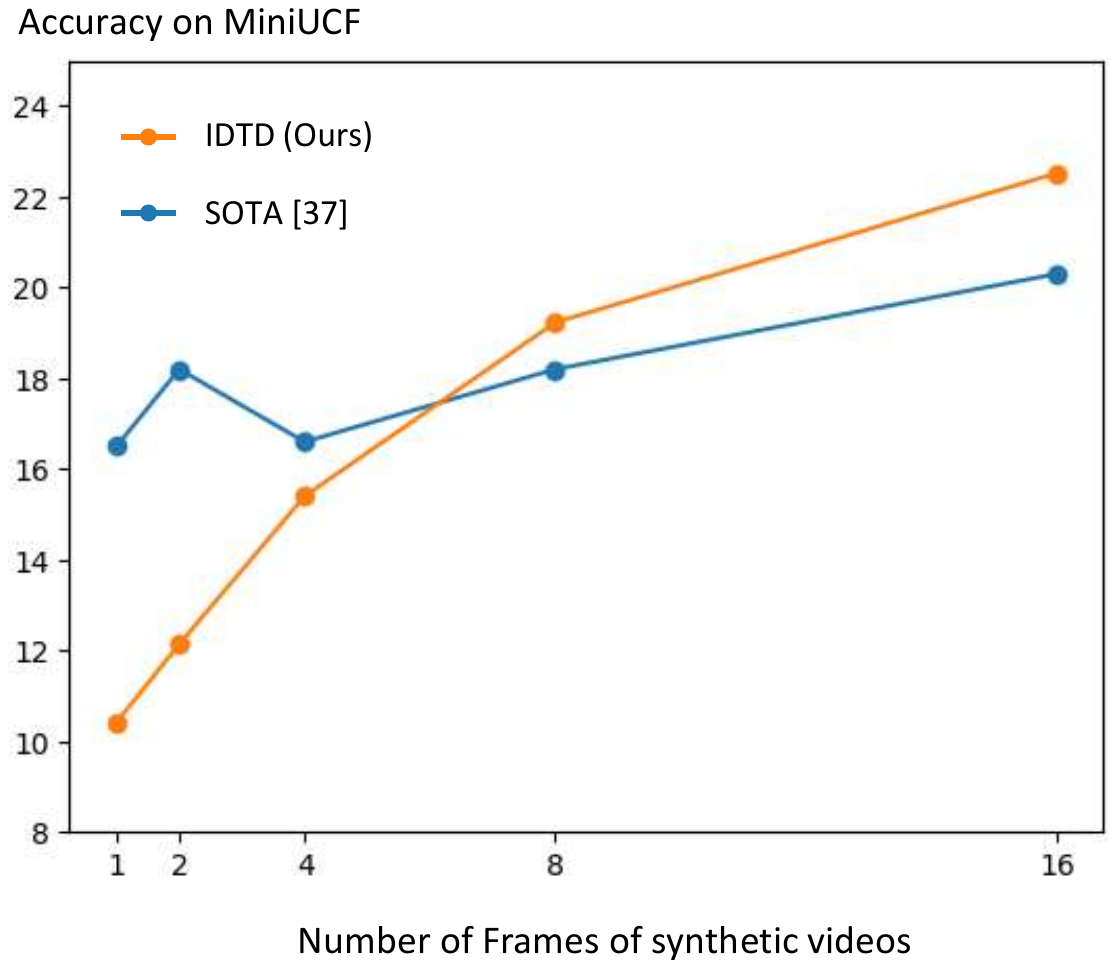}
    \caption{We compared the trend of performance as number of frames of synthetic videos changes. Ours shows a clear growth as the number of frames increases, while VDSD \cite{dd12} have no significant increase.}
    \label{fig:frame_num_abl}
    
\end{figure}

\begin{figure}[ht]
    
    \centering
    \includegraphics[width=0.45\textwidth,trim = 150 0 150 0]{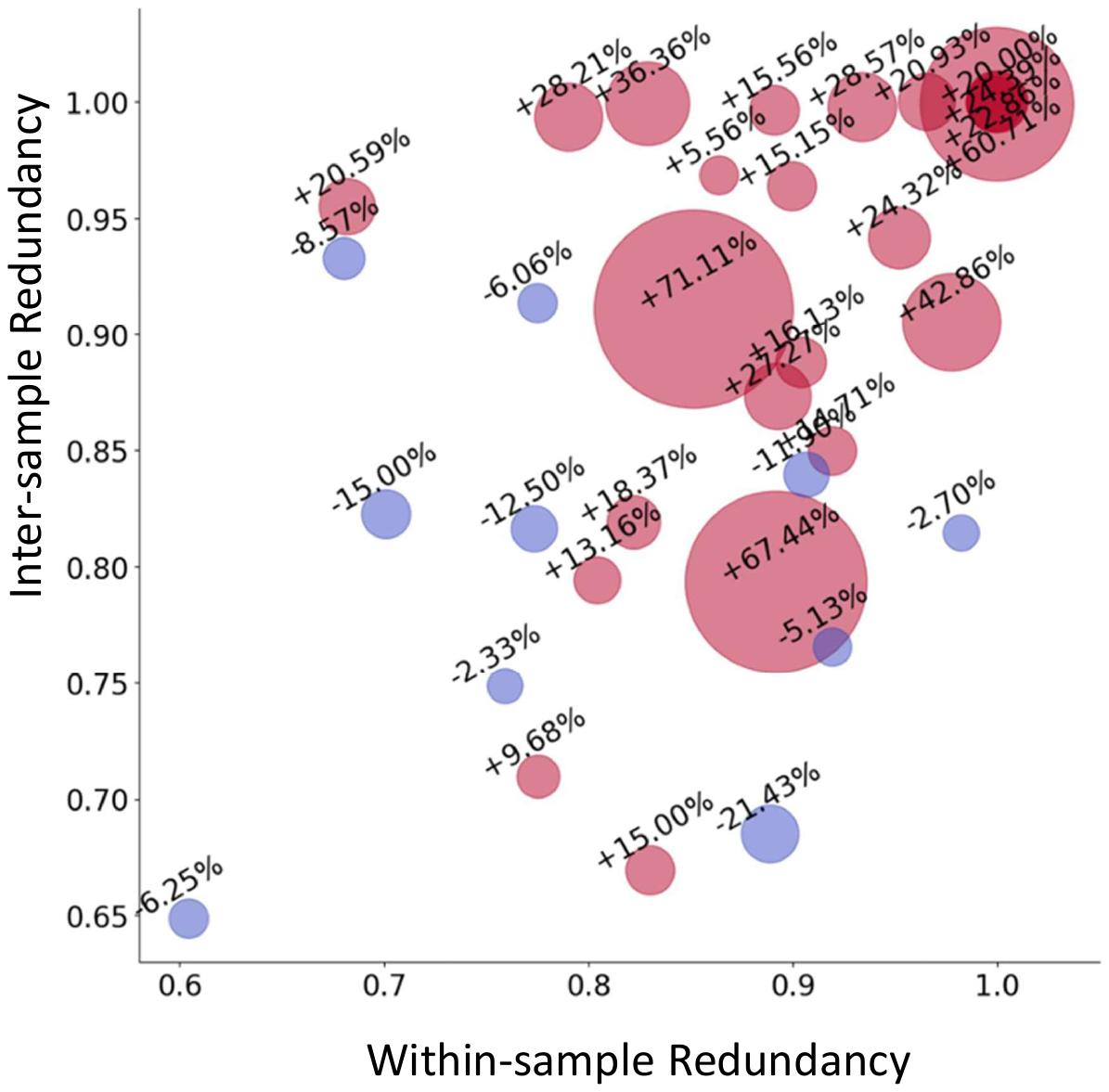}
    \caption{Redundancy Analysis. As shown in the diagram, our approach obtained higher performance gain where both inter-sample redundancy and within-sample redundancy are higher, indicating the effectiveness of joint optimization rather than naively discarding all temporal information followed by interpolation.}
    \label{fig:redun_anal}
    
\end{figure}

\textbf{Modules:} To verify the effectiveness of joint optimization of information diversification and temporal densification, We conducted ablation experiment on different modules as shown in Tab.\ref{tab:ablation1}, where Feature Selector aims to achieve an information diversifiction purpose, and the Temporal Fusor aims to achieve temporal densification purpose.

For Tab.\ref{tab:ablation1}-A, we randomly select $K$ real videos and naively do a temporal compression by temporal sampling and stitch them into a synthetic video. For Tab.\ref{tab:ablation1}-B, we conduct the experiment without Feature Pool or Feature Selectors, and we initialize the Diverse Segments with $K$ real videos and optimize the Diverse Segments and the Temporal Fusor. For Tab.\ref{tab:ablation1}-C, we optimzed only the Feature Pool and the Selectors and stitch the output of them (i.e. the Diverse Segments) into the synthetic videos.

From Tab.\ref{tab:ablation1}, we can tell that the joint optimzition of both the information diversification and the temporal densification,is a necessary condition to produce effective synthetic videos.

\begin{figure*}[ht]
    
    \centering
    \includegraphics[width=\textwidth, trim = 0 170 0 200]{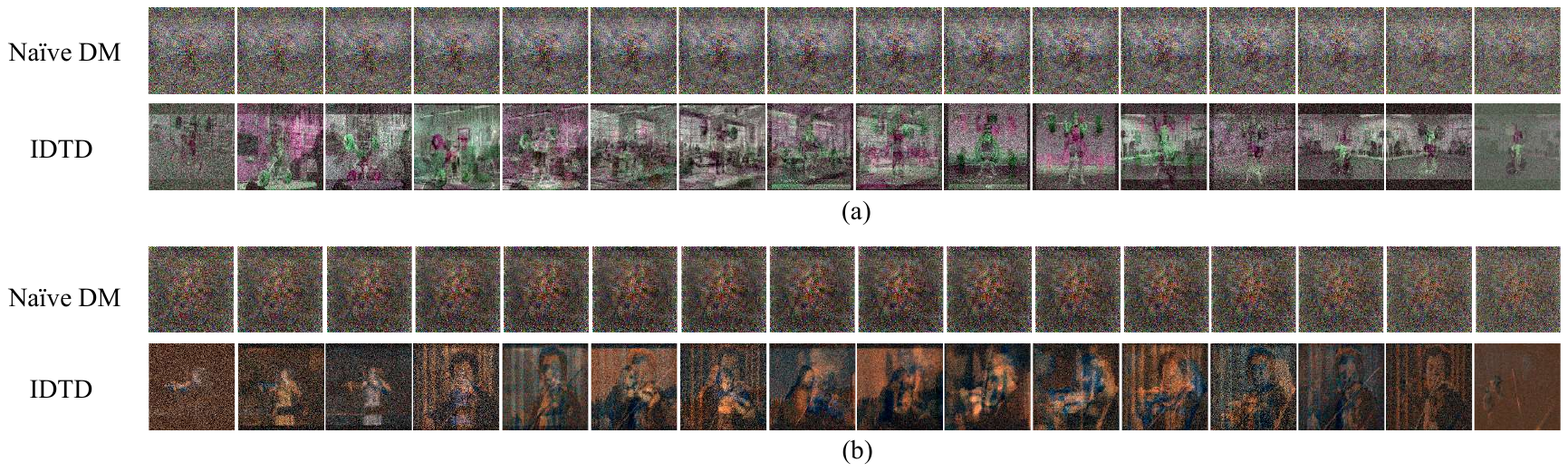}
    \caption{Qualitative Results compared between image-level approach (Distribution Matching) and our IDTD approach jointly optimizaing within-sample and inter-sample class redundancy. Our approach preserves much more temporal information diversity. Class (a) is CleanAndJerk and Class(b) is playing violin.}
    \label{fig:qual_result}
    
\end{figure*}

\textbf{Losses:} We further conduct our ablation study over the training objectives of our approach. The Feature Selector matching loss $\mathcal{L}_M$ enforces the Feature Pool and the Feature Selector to learn useful information and generate the feature segments similar with the feature of the real, while the Feature Selector diversity loss $\mathcal{L}_M$ pushed the feature of the Diverse Segments generated away from each other to diversely represent the real dataset.

We noticed that, even though our approach utilized an end-to-end training manner, naively applying the dataset distillation matching loss to the final output performs poorly as shown by Tab.\ref{tab:ablation2}-A. Afterthe diversity loss is applied as shown by  Tab.\ref{tab:ablation2}-B, it increases the performance but is still far below the full performance since the Feature Selector and Feature Pool do not get enough supervision signal from simply the matching loss applied to the Temporal Fusor. By applying the dataset distillation matching loss to both the Feature Selector and the Temporal Fusor as shown by Tab.\ref{tab:ablation2}-C, the performance is increases since the Feature Selector could implicitly obtain diverse features driven by a matching loss and also by sharing the same Feature Pool. We can see that the performance comes to its highest after all three training losses are applied as shown by the last row by Tab.\ref{tab:ablation2}.

\textbf{Temporal Size:} As shown in Fig.\ref{fig:frame_num_abl}, we conducted ablation on the frame number of one synthetic videos and compared to VDSD\cite{dd12}. The frame number of the real video is set as 16, and the synthetic frame number is changed. Our result shows a clear trend of performance increasing as the number of frames increases, while VDSD\cite{dd12} shows no obvious increase as they claimed. This is because our approach is able to preserve temporal information effectively and therefore is able to increase the performance as the number of the frames increases.

\textbf{Inter Class Analysis:} As shown in Fig.\ref{fig:redun_anal} a evaluation of performance improvement of our approach compared to Tab.\ref{tab:ablation1}-A is conducted. Each circle in the scatter plot is the performance improvement of a class, with the x-axis of within-sample redundancy and y-axis of inter-sample redundancy. The radial $r$ of the circle is positively correlated to the performance improvement. Namely, $r \propto (acc_{IDTD} - acc_{Tab.\ref{tab:ablation1}-A})$. The red color of the circle represents a positive performance gain, while the blue color represents a negative gain.

Retrieving feature with temporal information from a middle layer of the student model, we get temporal feature of the video $\mathcal{F}_t \in \mathcal{R}^{B \times t \times d}$, the temporal redundancy is defined as
$$R_{t} = \tanh(\frac{1}{\frac{1}{B}\sum_{b=1}^{B} var(\mathcal{F}_t(b))})$$ 
where $var(.)$ is the variance along the first dimension of any variable. Retrieving a video's feature encoded by the second last layer of the student model $\mathcal{F}_{IC} \in \mathcal{R}^{B \times \Tilde{d}}$, the inter-sample redundancy defined as

$$R_{IC} = \tanh(\frac{1}{\frac{1}{B}var(\mathcal{F}_{IC})})$$

Therefore in this way, we calculate redundancy in both dimensions by measuring the reciprocal of variance and normalizing the value to $[0,1]$.

As shown in Fig.\ref{fig:redun_anal}, as the within-sample redundancy and inter-sample redundancy increases, the performance gain in the classes significantly increases. We can also notice that, the performance gain achieved becomes the largest when the inter-sample redundancy and the within-sample redundancy is both high.

\subsection{Qualitative Results}

We compare the qualitative results with an image-level dataset distillation approach (Distribution Matching\cite{dd10}). As shown in Fig.\ref{fig:qual_result}, the results of Distribution Matching approach and our IDTD approach are shown. Our synthetic videos are significantly more diverse along the temporal axis and contains more temporal dynamic information than the image level dataset distillation or than VDSD\cite{dd12} which generated the synthetic video by interpolating a distilled static image.

\section{Conclusion}

In conclusion, in this work we are the first to study the problem of Video Set Distillation, and we proposed a approach to jointly reduce both the inter-sample redundancy and within-sample redundancy in an end-to-end manner. Our approach leveraged a diversity loss to produce more diverse and representative features for synthetic videos. A Temporal Fusor is applied to densify temporal information while preserving the diversity. Our approach achieved the SOTA on most of the dataset evaluations with a more realistic and temporally diverse synthetic videos. 

{
    \small
    \bibliographystyle{ieeenat_fullname}

\begin{thebibliography}{26}
\providecommand{\natexlab}[1]{#1}
\providecommand{\url}[1]{\texttt{#1}}
\expandafter\ifx\csname urlstyle\endcsname\relax
  \providecommand{\doi}[1]{doi: #1}\else
  \providecommand{\doi}{doi: \begingroup \urlstyle{rm}\Url}\fi

\bibitem[Carreira and Zisserman(2017)]{vidrec6}
Joao Carreira and Andrew Zisserman.
\newblock Quo vadis and action recognition? a new model and the kinetics dataset.
\newblock In \emph{CVPR}, 2017.

\bibitem[Cazenavette et~al.(2022)Cazenavette, Wang, Antonio~Torralba, , and Zhu]{dd2}
George Cazenavette, Tongzhou Wang, Alexei A~Efros Antonio~Torralba, , and Jun-Yan Zhu.
\newblock Dataset distillation by matching training trajectories.
\newblock In \emph{CVPR}, 2022.

\bibitem[Cui et~al.(2023)Cui, Wang, Si, , and Hsieh]{dd4}
Justin Cui, Ruochen Wang, Si Si, , and Cho-Jui Hsieh.
\newblock Scaling up dataset distillation to imagenet-1k with constant memory.
\newblock In \emph{IMCL}, 2023.

\bibitem[Dirfaux(2000)]{naive_met1}
F. Dirfaux.
\newblock Key frame selection to represent a video.
\newblock In \emph{Proceedings 2000 International Conference on Image Processing}, 2000.

\bibitem[Feichtenhofer et~al.(2016)Feichtenhofer, Pinz, , and Wildes.]{vidrec3}
Christoph Feichtenhofer, Axel Pinz, , and Richard~P Wildes.
\newblock Spatiotemporal residual networks for video action recognition.
\newblock In \emph{NeurIPS}, 2016.

\bibitem[Feichtenhofer et~al.(2019)Feichtenhofer, Fan, Malik, and He]{vidrec2}
Christoph Feichtenhofer, Haoqi Fan, Jitendra Malik, and Kaiming He.
\newblock Slowfast networks for video recognition.
\newblock In \emph{ICCV}, 2019.

\bibitem[Goyal et~al.(2017)Goyal, Kahou, ski, nska, Westphal, Kim, Haenel, Fruend, Yianilos, Mueller-Freitag, Hoppe, Thurau, Bax, and Memisevic]{dst3}
Raghav Goyal, Samira~Ebrahimi Kahou, Vincent~Michal ski, Joanna~Materzy´ nska, Susanne Westphal, Heuna Kim, Valentin Haenel, Ingo Fruend, Peter Yianilos, Moritz Mueller-Freitag, Florian Hoppe, Christian Thurau, Ingo Bax, and Roland Memisevic.
\newblock The ”something something” video database for learning and evaluating visual common sense.
\newblock In \emph{ICCV}, 2017.

\bibitem[He et~al.(2016)He, Zhang, Ren, and Sun]{vidrec4}
Kaiming He, Xiangyu Zhang, Shaoqing Ren, and Jian Sun.
\newblock Deep residual learning for image recognition.
\newblock In \emph{CVPR}, 2016.

\bibitem[Karpathy et~al.(2014)Karpathy, Toderici, Shetty, Leung, Sukthankar, and Fei-Fei]{vidrec1}
Andrej Karpathy, George Toderici, Sanketh Shetty, Thomas Leung, Rahul Sukthankar, and Li Fei-Fei.
\newblock Large-scale video classification with convolutional neural networks.
\newblock In \emph{CVPR}, 2014.

\bibitem[Kuehne et~al.(2011)Kuehne, Jhuang, Garrote, Poggio, , and Serre]{dst2}
H. Kuehne, H. Jhuang, E. Garrote, T. Poggio, , and T. Serre.
\newblock Hmdb: A large video database for human motion recogni tion.
\newblock In \emph{ICCV}, 2011.

\bibitem[Li and Li.(2021)]{dd5}
Yongqi Li and Wenjie Li.
\newblock Data distillation for text classification.
\newblock In \emph{arXiv preprint arXiv:2104.08448}, 2021.

\bibitem[MaxWelling(2009)]{exp1}
MaxWelling.
\newblock Herding dynamical weights to learn.
\newblock In \emph{ICML}, 2009.

\bibitem[Sener and Savarese(2018)]{exp2}
Ozan Sener and Silvio Savarese.
\newblock Active learning for con volutional neural networks: A core-set approach.
\newblock In \emph{ICLR}, 2018.

\bibitem[Soomro et~al.(2012)Soomro, Zamir, , and Shah]{dst1}
Khurram Soomro, Amir~Roshan Zamir, , and Mubarak Shah.
\newblock Ucf101: A dataset of 101 human actions classes from videos in the wild.
\newblock In \emph{CoRR}, 2012.

\bibitem[Sun et~al.(2024)Sun, Shi, Yu, and Lin]{Diversity_4}
Peng Sun, Bei Shi, Daiwei Yu, and Tao Lin.
\newblock On the diversity and realism of distilled dataset: Anefficient dataset distillation paradigm.
\newblock In \emph{CVPR}, 2024.

\bibitem[Tran et~al.(2015)Tran, Bourdev, Fergus, Torresani, and Paluri]{vidrec5}
Du Tran, Lubomir Bourdev, Rob Fergus, Lorenzo Torresani, and Manohar Paluri.
\newblock Learning spatiotemporal features with 3d convolutional networks.
\newblock In \emph{ICCV}, 2015.

\bibitem[Wang et~al.(2022)Wang, Zhao, Peng, Zhu, Shuo~Yang, Huang, Bilen, Wang, , and You]{dd7}
Kai Wang, Bo Zhao, Xiangyu Peng, Zheng Zhu, Shuo~Wang Shuo~Yang, Guan Huang, Hakan Bilen, Xinchao Wang, , and Yang You.
\newblock Cafe: Learning to condense dataset by aligning features.
\newblock In \emph{CVPR}, 2022.

\bibitem[Wang et~al.(2018)Wang, Zhu, Torralba, , and Efros]{dd1}
Tongzhou Wang, Jun-Yan Zhu, Antonio Torralba, , and Alexei~A Efros.
\newblock Dataset distillation.
\newblock In \emph{arXiv preprint arXiv:1811.10959}, 2018.

\bibitem[Wang et~al.(2024)Wang, Xu, Lu, and Li]{dd12}
Ziyu Wang, Yue Xu, Cewu Lu, and Yong-Lu Li.
\newblock Dancing with still images: Video distillation via static-dynamic disentanglement.
\newblock In \emph{CVPR}, 2024.

\bibitem[Wu et~al.(2023)Wu, Deng, , and Russakovsky]{dd8}
Xindi Wu, Zhiwei Deng, , and Olga Russakovsky.
\newblock Multimodal dataset distillation for image-text retrieval.
\newblock In \emph{arXiv preprint arXiv:2308.07545}, 2023.

\bibitem[Yang et~al.(2000)Yang, HanyuPeng, MinXu, MingmingSun, and PingLi]{naive_met2}
Shuo Yang, ZekeXie HanyuPeng, MinXu, MingmingSun, and PingLi.
\newblock Dataset pruning: Reducing training data by examining generalization influence.
\newblock In \emph{Proceedings 2000 International Conference on Image Processing}, 2000.

\bibitem[Yin et~al.(2023)Yin, Xing, and Shen]{dd4_1}
Zeyuan Yin, Eric Xing, and Zhiqiang Shen.
\newblock Squeeze and recover and relabel: Dataset condensation at imagenet scale from a new perspective.
\newblock In \emph{NeurIPS}, 2023.

\bibitem[Zhao and Bilen(2023)]{dd10}
Bo Zhao and Hakan Bilen.
\newblock Dataset condensation with distribution matching.
\newblock In \emph{WACV}, 2023.

\bibitem[Zhao et~al.(2021)Zhao, Mopuri, , and Bilen]{dd3}
Bo Zhao, Konda~Reddy Mopuri, , and Hakan Bilen.
\newblock Dataset condensation with gradient matching.
\newblock In \emph{ICLR}, 2021.

\bibitem[Zhao et~al.(2023)Zhao, Li, Qin, and Yu]{dd13}
Ganlong Zhao, Guanbin Li, Yipeng Qin, and Yizhou Yu.
\newblock Improved distribution matching for dataset condensation.
\newblock In \emph{CVPR}, 2023.

\bibitem[Zhou et~al.(2022)Zhou, Nezhadarya, , and Ba]{dd11}
Yongchao Zhou, Ehsan Nezhadarya, , and Jimmy Ba.
\newblock Dataset distillation using neural feature regression.
\newblock In \emph{NeurIPS}, 2022.

\end{thebibliography}

}


\end{document}